\documentclass[10pt,twocolumn,letterpaper]{article}

\usepackage[pagenumbers]{cvpr} %

\usepackage[accsupp]{axessibility}
\usepackage{graphicx}
\usepackage{amsmath}
\usepackage{amssymb}
\usepackage{bm}
\usepackage{booktabs}
\usepackage{siunitx}
\usepackage{caption}
\usepackage{subcaption}

\usepackage[pagebackref,breaklinks,colorlinks]{hyperref}

\usepackage[capitalize]{cleveref}
\crefname{section}{Sec.}{Secs.}
\Crefname{section}{Section}{Sections}
\Crefname{table}{Table}{Tables}
\crefname{table}{Tab.}{Tabs.}

\def\cvprPaperID{32}
\def\confName{CVPR}
\def\confYear{2023}
\begin{document}

\title{X-maps: Direct Depth Lookup for Event-based Structured Light Systems}

\author{
Wieland Morgenstern\textsuperscript{1}
\quad Niklas Gard\textsuperscript{1,2}
\quad Simon Baumann\textsuperscript{1}
\quad Anna Hilsmann\textsuperscript{1} 
\quad Peter Eisert\textsuperscript{1,2} 
\\[0.2cm]
\textsuperscript{1} Fraunhofer Heinrich Hertz Institute, HHI \\ \textsuperscript{2} Humboldt University of Berlin
\\[0.2cm]
{\tt\small \{first\}.\{last\}@hhi.fraunhofer.de}
}
\maketitle

\begin{abstract}
We present a new approach to direct depth estimation for Spatial Augmented Reality (SAR) applications using event cameras. These dynamic vision sensors are a great fit to be paired with laser projectors for depth estimation in a structured light approach.
Our key contributions involve a conversion of the projector time map into a rectified \mbox{X-map}, capturing x-axis correspondences for incoming events and enabling direct disparity lookup without any additional search. Compared to previous implementations, this significantly simplifies depth estimation, making it more efficient, while the accuracy is similar to the time map-based process. Moreover, we compensate non-linear temporal behavior of cheap laser projectors by a simple time map calibration, resulting in improved performance and increased depth estimation accuracy.
Since depth estimation is executed by two lookups only, it can be executed almost instantly (less than 3~ms per frame with a Python implementation) for incoming events. This allows for real-time interactivity and responsiveness, which makes our approach especially suitable for SAR experiences where low latency, high frame rates and direct feedback are crucial. 
We present valuable insights gained into data transformed into X-maps and evaluate our depth from disparity estimation against the state of the art time map-based results. Additional results and code are available on our project page: \href{https://fraunhoferhhi.github.io/X-maps/}{fraunhoferhhi.github.io/X-maps}.
\end{abstract}

\section{Introduction}
\label{sec:intro}

Spatial Augmented Reality (SAR) has emerged as a promising technology that seamlessly merges digital information with the physical world. Also known as projection-based augmented reality, SAR focuses on overlaying virtual content onto the physical environment \cite{Raskar2001}. This offers the potential to create immersive, interactive, and engaging experiences across various domains: SAR is utilized in entertainment \cite{Bermano2017}, industrial applications \cite{Andersen16, Doshi2017}, advertising \cite{Park2015}, cultural heritage\cite{Ridel2014}, and healthcare \cite{Chae2018}.

\begin{figure}[t]
  \centering
  \begin{subfigure}{0.3\columnwidth}
    \includegraphics[height=1.9in]{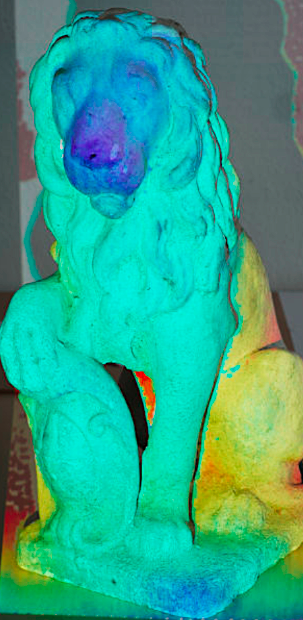}
    \caption{} 
  \end{subfigure}
  \begin{subfigure}{0.34\columnwidth}
    \includegraphics[height=1.9in]{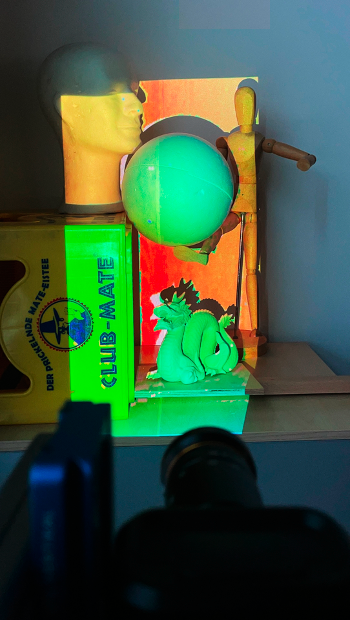}
    \caption{} 
  \end{subfigure}
    \begin{subfigure}{0.3\columnwidth}
    \includegraphics[height=1.9in]{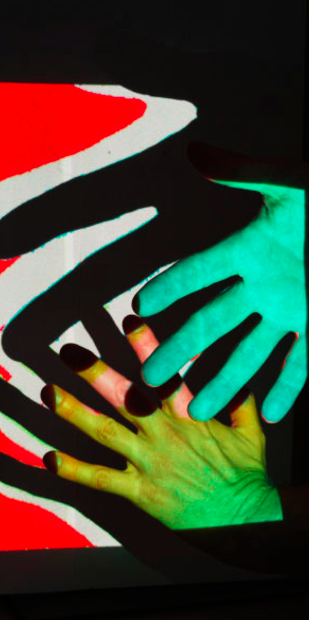}
    \caption{} 
  \end{subfigure}
   \caption{Through the use of X-maps, we can establish real-time Spatial Augmented Reality (SAR) applications, using an event camera and laser projector system. We calculate depth from the projection with minimal computational effort at high frame rates. Our demonstrator projects the color coded depth back into the scene. Here, (a) is a static scene, (b) shows the projector and camera looking onto the scene with the projection in the background, and (c) is demonstrating depth estimation on moving objects.}
   \label{fig:comb_scene_shots}
\end{figure}

Accurate estimation of depth in real-time is essential for achieving compelling and responsive augmented reality experiences. In this paper, we introduce a novel approach to ultra low latency depth estimation for SAR applications by leveraging event cameras and structured light projectors.

Event cameras are a groundbreaking technology that offers significant advantages in terms of speed and robustness compared to traditional frame-based cameras \cite{gallego2020event}. By measuring changes in pixel intensity, event cameras can capture motion and brightness variations with high temporal resolution and dynamic range. This makes them ideal candidates for integration with structured light projectors in SAR systems.

Our approach builds on the findings of previous methods \cite{Matsuda2015, Muglikar2021}, which combine structured light technology and event cameras to address the inherent limitations of conventional structured light systems. However, our method does not require an external hardware synchronization device between projector and camera to identify individual frame starts and ends, leading to a simpler setup and potentially more cost-effective projector options. Furthermore, the real-time capability of our system is a key enabler for SAR applications. The ability to perform depth estimation and reconstruction with minimal computational effort opens up numerous use cases, such as reprojecting depth, projecting textures onto objects or planes, and even enabling partial projection in challenging scenarios.

We make the following contributions:
\begin{enumerate}
    \item We transform the projector time map into a rectified X-map that captures the x-axis correspondences for incoming events, enabling the direct lookup of the disparity.
    \item To model the nonlinear temporal behavior of consumer micro-electro-mechanical systems (MEMS) laser scanners, we calibrate the time map and improve depth estimation errors by pointing the projector-event-camera-system at a planar surface.
    \item As depth estimation is performed with low computational effort, our system is suitable for spatial augmented reality applications that require high frame rates and direct feedback. We demonstrate a proof of concept that is, to our knowledge, the first real-time event-based depth estimation system with integrated content projection.
\end{enumerate}

\section{Related Work}
\label{sec:related_work}

\begin{figure}[t]
  \centering
   \includegraphics[width=\linewidth]{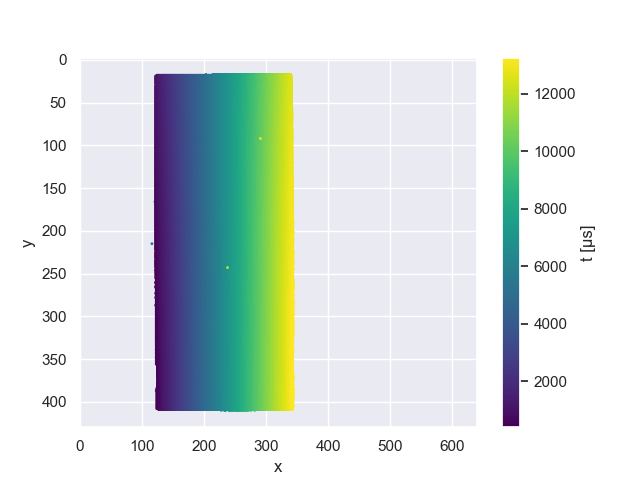}

   \caption{A \textit{time map} of recorded events for a single frame projected by a laser projector onto a plane. Matching time entries of the map along epipolar lines with an idealized projector time map to compute scene disparity is computationally expensive \cite{Muglikar2021}.}
   \label{fig:time-map-full}
\end{figure}

Commercial structured light systems, such as the Microsoft Kinect \cite{KinectV1} and Intel RealSense\cite{IntelRealSense}, have brought depth sensing technology to mainstream applications like gaming, robotics, and computer vision. These systems usually provide at most 30 frames per second (fps), which can result in delayed or inaccurate depth information, hindering real-time applications and smooth user experiences. Additionally, the structured light projection used in these systems targets depth estimation and does not convey useful information or visuals, limiting their potential in applications like Spatial Augmented Reality (SAR) that require informative projections. There are approaches, which use projected content, captured by a frame-based camera, directly to infer scene parameters, e.g.~for object \cite{Gard2019} or projection plane tracking \cite{Kagami2019}. While they deliver very accurate mapping results \cite{Gard2019} and are capable of high frame rates \cite{Kagami2019}, they assume model knowledge to project on specific objects, but do not estimate the depth for the entire projection area. It has been shown that many classical image-based methods that do not pay separate attention to projection are distracted by the projected content in the camera image \cite{Zheng2013}.

Previously, several works have looked at combining event camera sensors with laser projection. After the pioneering method by Brandli \etal \cite{Brandli2014} combined a dynamic vision sensor with a laser line projector, methods that scan the entire image quickly emerged. %
Motion Contrast 3D (MC3D) \cite{Matsuda2015} introduces the concept of merging single-shot structured light techniques with event-based cameras and addresses the trade-offs between resolution, robustness, and speed in structured light systems. MC3D employs a projector that scans scenes using a single laser beam in a raster pattern. At high sampling rates, MC3D's correspondence search amplifies noise in the event timestamps, resulting in noisy and patchy stereo correspondences.

\begin{figure}[t]
  \centering
   \includegraphics[width=\linewidth]{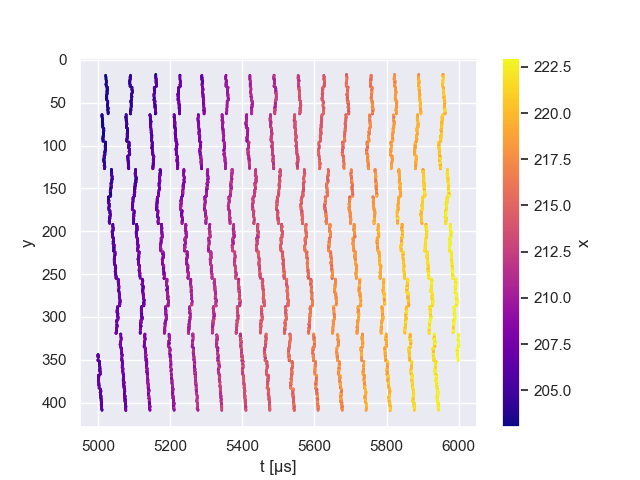}
   \caption{When we plot incoming events of a projected frame with their y coordinate over time, we can clearly separate them into columns, grouped by their time stamps.}
   \label{fig:x-map-zoom-x}
\end{figure}

Building on MC3D's groundwork, Event-based Structured Light (ESL) \cite{Muglikar2021} introduces time maps to establish the temporal projector-camera correspondences. A camera time map is illustrated in Figure \ref{fig:time-map-full}. After initializing depth maps through an epipolar disparity search in rectified projector time maps, an extra processing step optimizes consistency within a local window around each pixel. The optimization step reduces the influence of event jitter, but is computationally very expensive. The performance is compared to MC3D and Semi-Global Matching (SGM) \cite{Hirschmuller2007} algorithms: ESL surpasses both MC3D and SGM in static scenes, demonstrating lower error values and superior noise suppression against the baseline. Due to the complex correspondence search and optimization process, is not real-time capable.

Foveated rendering in VR headsets is adapted for depth sensing in \cite{Muglikar2021b}, where the authors develop a foveating depth sensing demonstrator. Static scene parts are sparsely scanned, while moving parts are densely scanned, and a neural network completes depth from sparse samples. This approach reduces the scanned area to approximately 10\%, decreasing power consumption and time stamp jitter in event data, as fewer events need processing on the chip.

The authors of \cite{Xueyan2021} combine an event camera with a DLP projector and demonstrate that scan rates of up to 1000 fps might be technically feasible. They project a fixed pattern, similar to the pattern of a Kinect sensor, and do not utilize the time domain of the projector (despite ON/OFF), so inferring depth with other projected content is not possible with their system, ruling out its use for spatial augmented reality.

\section{Direct Depth Lookup from Events}

\begin{figure}[t]
  \centering
   \includegraphics[width=\linewidth]{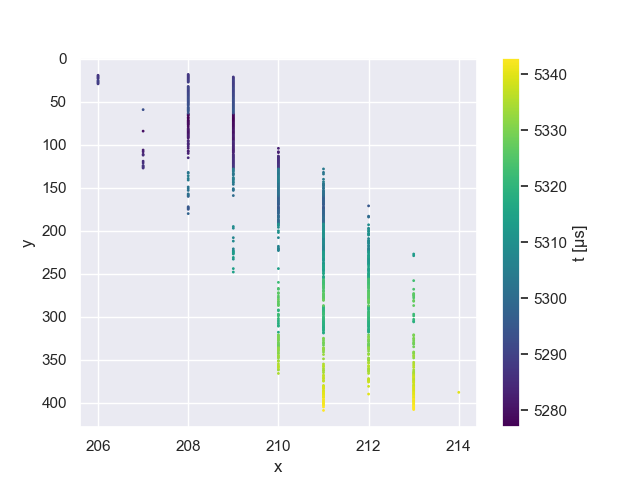}

   \caption{This figure shows all events of a single column from Figure \ref{fig:x-map-zoom-x}. Even though we are projecting onto a plane, we can see that events of the same temporal local group have jitter with 2-3 pixels on the x-axis. The jitter will lead to events overlaying with those of the neighboring columns. Thus it is not possible to clearly distinguish projected scan lines in the y/x view, while it is possible in the y/t view in Figure \ref{fig:x-map-zoom-x}.}
   \label{fig:time-map-col75}
\end{figure}

In this section, we introduce our approach to estimate the scene depth from the projected image and the resulting events in the camera.

Event camera sensors produce asynchronous events independently for every pixel \cite{gallego2020event}. The incoming event stream delivers camera coordinates $(\bm{x}, \bm{y})$, a timestamp $\bm{t}$ and a polarity $\bm{p}$ for each event. Storing the time stamps of events in an image the size of the camera's resolution results in a time map of $(x,y) \longmapsto t$, displayed in Figure \ref{fig:time-map-full}. A rectified time map is used in ESL \cite{Muglikar2021} to match the recorded time in the camera's time map to the ideal time in a synthesized projector time map, by searching along epipolar lines to minimize the difference in $t$. As we target real-time AR applications, we want to avoid an exhaustive disparity search. While it is feasible to do the initial disparity search on a high-performing GPU, there is no clear path to speed up the following per-pixel optimization process. We instead propose a solution to directly retrieve the disparity, which is possible with a simple and fast CPU implementation.

Like in ESL, we have tilted our projector 90 degrees, so that the projector rows scan from the bottom to the top, and the projector columns scan from left to right in the scene. Thus, the larger time steps align with the axis of the disparity. Through the projector's rotation, one projector scan row now ideally forms an almost vertical line in the scene.

\begin{figure}[t]
  \centering
   \includegraphics[width=\linewidth]{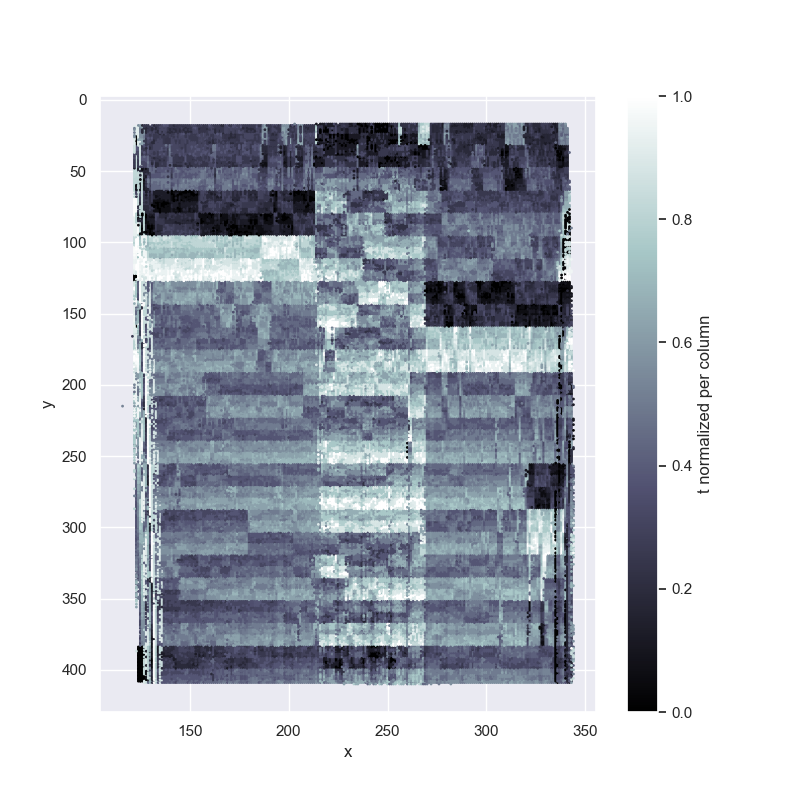}

   \caption{Column event time in a time map for a projected frame. Incoming events were grouped by their time (splitting the event stream into the columns seen in Figure \ref{fig:x-map-zoom-x}). Per column, the time stamp of each event was then normalized to the range $[0,1]$. In an ideal recording, we would expect a smooth gradient from the bottom to the top of the image, as the projector is scanning each line vertically. In practice, we have different effects of the read-out mechanism of the camera adding up to heavy local time stamp jitter.}
   \label{fig:projector_rowtime}
\end{figure}

We can track the events these columns create in each projected frame, by plotting incoming events with their $y$ coordinate over the time, as in Figure \ref{fig:x-map-zoom-x}. The timestamps within a projector row do not contain significant information, as there are too many sources of local signal distortions: readout noise, neighboring pixels getting triggered at the same time from a single pulse of laser, projector/camera resolution mismatch. The results of these effect can be seen in the single line plot in Figure \ref{fig:time-map-col75}, and the full frame time normalized plot in Figure \ref{fig:projector_rowtime}. A good discussion of the different sources of noise within the camera timestamps can be found in \cite{Muglikar2021b}. The authors discuss how transistor noise, parasitic photo currents, and the arbitrated architecture of the sensor lead to deviations from the ideal time stamp generation.

As the time stamp is not sufficiently precise locally, a direct map from the time stamp alone into the projector's view is not possible. We could however determine the epipolar line from our coordinates $(x, y)$ and intersect this with the projector row, determined from the time stamp. This works under the assumption that the projector behaves linearly over time. However, as we show in Figure \ref{fig:time_map_calib}, the projector may not scan at a constant speed.

We are thus looking for a method, where we can directly compute the disparity (for real-time capability), and work with projectors that show non-linear time behavior.

Our idea is to provide a lookup table for the disparity values to avoid the disparity search required when using time maps. We store the projector $x$ coordinates over $y$ and the time $t$:
\begin{equation}
 (y,t) \longmapsto x   
\end{equation}

The time axis can be freely discretized. We choose to make it the projected width $w$ of the projector, which should allow different scan lines to map to different time columns. In practice, the projector resolution is often higher than that of the camera sensor and covering about a third of the area, thus the event camera will not see single projector rows.

\subsection{Creating a reference X-map}
\label{sec:projector-reference-x-map}

As the lookup reference, we require a projector X-map. We will construct this map from the projector time map, which allows us to use calibrated time maps, correcting non-linear temporal projector behavior (see Section \ref{sec:time_map_calibration}). We want the projector X-map to be filled completely within the projector's bounds, so we have values for all possible measurements from the camera; we do not know beforehand, which $(y,t)$ combinations we might see. To retrieve matching x values from the time map for the projector \mbox{X-map}, we use a search along the epipolar lines akin to the disparity search from ESL, to be able to fill out our complete rectified X-map. This needs to be done only once, after calibrating the time map.

To fill the projector X-map $(y, t) \longmapsto x$ at rectified projector coordinate $y_{p_r}$ and time $t$, we find the rectified coordinate $x_r$ that minimizes the difference between the value retrieved from the calibrated time map $m$. We do not consider x values that are further than two rows away, by comparing the time difference to projector width $w$:

\begin{equation}
x(y_{p_r}, t) = \min_{x_r} \begin{cases}
t_d = | t - m(x_r, y_{p_r}) |,& \text{if } t_d \leq \frac{2}{w}\\
\inf,              & \text{otherwise}
\end{cases}
\end{equation}

\begin{figure}[t]
  \centering
   \includegraphics[width=\linewidth]{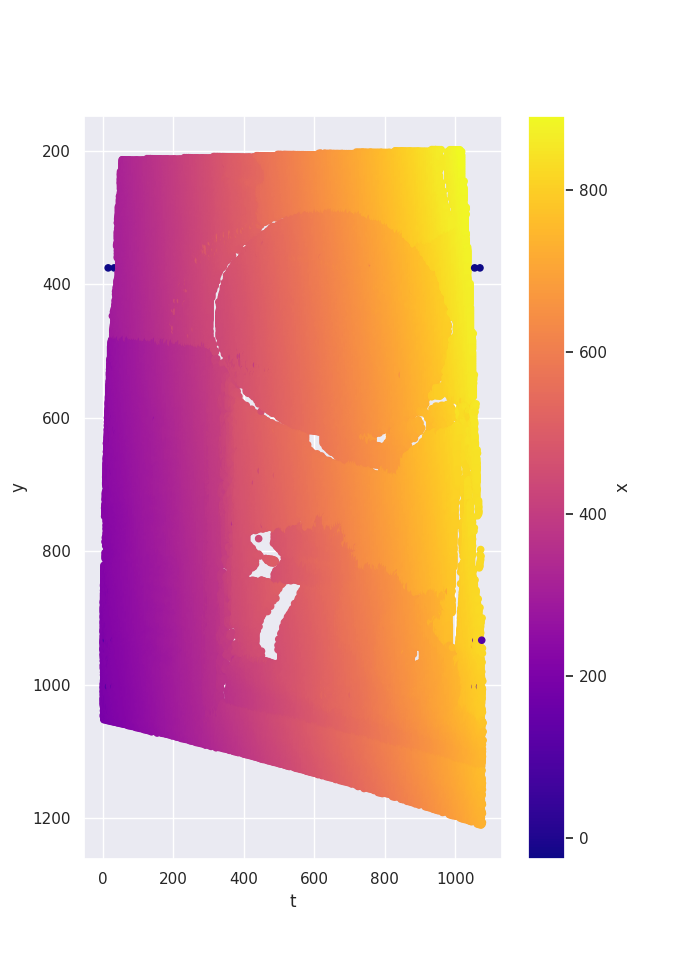}

   \caption{A rectified camera X-map for a single projected frame. It shows the scene of Figure \ref{fig:comb_scene_shots} (b). The X-map is the product of flattening the spatio-temporal event cuboid of dimensions $(x,y,t)$ into a 2D image of $(y,t) \longmapsto x$. The idea of the time map is similar, but applies the concept to a different face of the cuboid (mapping $(x,y) \longmapsto t$). The X-map forms the basis for our method. As x values are the values encoded in the map, subtracting a rectified camera X-map from an idealized projector X-map yields the disparity value.}
   \label{fig:x-map-full}
\end{figure}

\subsection{Direct disparity lookup in X-maps}
\label{sec:disparity-look-up}

Given the dense rectified projector X-map, we can lookup the disparity value for each incoming event. Event coordinates $(x,y)$ are rectified into $x_{c_r}$ and $y_{c_r}$. The event time $t$ is scaled and discretized to the width of the X-map.

\begin{equation}
x_{p_r} = x(y_{c_r}, t)
\end{equation}
\begin{equation}
d = x_{p_r} - x_{c_r}
\end{equation}

Events that have coordinates which map into an undefined area of the X-map or which result in a disparity value $d < 0$  are discarded. As both operations (rectification and x value retrieval) are lookups, computation of the disparity is very efficient. The camera matrices are known from the calibration process, so we can now compute the depth and 3D coordinates for each event from our retrieved disparity values. To illustrate the process, we have created a camera X-map in Figure \ref{fig:x-map-full}. As we can process each event individually, there is no need to construct a camera X-map for the actual implementation.

\subsection{Time map calibration}
\label{sec:time_map_calibration}

The ideal time map for the projector assumes a constant laser speed across each pixel and an instantaneous jump between line ends. However, this fails to account for the projector's non-linear scan time, jitter introduced by the camera readout behavior, and deviations from the assumed raster pattern. To generate a more accurate reference time map, a calibration of the projector scanning behavior is necessary. Our solution is to project a white image onto a stationary plane parallel to the projector and camera image planes, obtaining a time map that directly corresponds to the actual projector time map as seen by the camera.

To accommodate the camera's different resolution, we generate multiple time maps, normalize them to the range (0, 1), and compute the pixel-wise average, filtering out noise and jitter. Then, we create a binary map, identify the projected frame's corners, and calculate the projective transformation from the irregular rectangle to a 2D map with the projector's width and height. The calibrated projector time map, flattened to one dimension, reveals the non-linearity in time, as the projector starts slower than the ideal curve but finishes faster.

Since the projected frame is not aligned with the camera lines and columns, most pixels in the resulting time map are interpolated and warped. To enable differentiation between projector lines, we interpolate missing values linearly between two actual lines. Despite these adjustments, time\-stamps do not increase monotonically, as seen in the ideal time map or a perfectly accurate time map. The calibrated time map can mitigate unexpected temporal behavior, even without complete knowledge of the projector's inner workings.

\section{Demonstrator Setup}

\label{sec:setup}

We build our demonstrator using the Nebra Anybeam MEMS Laser Projector \cite{nebra_anybeam}. This projector features three laser diodes scanning a resolution of $1280 \times 720$ px, a fixed refresh rate of $\SI{60}{\hertz}$ and it specifies a brightness of 30 ANSI lumens. The frame rate and resolution of the projector yield a rate of $ 1 / (720 \times \SI{60}{\hertz}) = \SI{23.2}{\us}$ per vertical projected line, which is well above the time stamp resolution of the event camera, but potentially shorter than the refractory period (the duration after firing, until the pixel can generate a new event).

For the event camera, a Prophesee Evaluation Kit 1 (EVK1) with a Gen 3.0 sensor is used. The dynamic vision sensor has a resolution of $640 \times 480$ px with a $\SI{15}{\micro\meter}$ pixel pitch and provides contrast detections only. The sensor has a dynamic range of greater than $\SI{120}{\decibel}$, a typical latency of $\SI{200}{\us}$ and timestamps the events with microsecond precision

\subsection{Camera-projector-calibration}

Camera-projector systems can be calibrated with established methods such as \cite{Moreno2012}. A camera calibration framework tailored to event cameras is presented in \cite{how-to-calibrate}, which converts event streams into intensity images with \cite{e2vid}, and thus does not require a blinking LED screen.

For our demonstrator, we calibrated our system with a custom calibration pipeline that first calibrates camera intrinsics with a checkerboard on a monitor alternating with all-white at 30 Hz. Then, we used a white diffusion paper glued to the same monitor to perform intrinsic projector and extrinsic projector camera calibration. The diffusion paper improved the monitor's ability to reflect the image from the projector, while maintaining the sharpness of the displayed checkerboard. Our approach was usable for system calibration, but the reprojection errors did not improve over conventional calibration methods, perhaps because of the error introduced by the diffusion paper not sharing the exact same plane as the monitor.

\subsection{Frame triggers}
\label{sec:frame_triggers}

The creation of time maps for event cameras involves identifying the start and end times within the event stream corresponding to a projected frame. While some event cameras offer external trigger connections, this method requires additional hardware. It also removes the possibility to use cheap consumer laser projectors, as they usually do not offer trigger hardware.
We analyze batches of events spanning more than $\SI{16}{\ms}$, to ensure that at least one frame gap is present at $\SI{60}{\hertz}$. We have observed our projector to continually scan over the scene for around \SI{13}{\ms}, followed by a reset phase of around \SI{3}{\ms}.

We look for sequences of events that have a small difference in time stamps $\Delta t$ between consecutive events (usually $\SI{0}{\us}$ or $\SI{1}{\us}$ while scanning). A frame is identified by a sequence of events that spans at least half a frame length (\SI{8}{\ms}), where no $\Delta t$ is larger than a threshold. For our projector, the threshold was chosen empirically at $\SI{40}{\us}$, which is around the time to scan two lines (see Section \ref{sec:setup}). Although the dataset in ESL was recorded with a hardware trigger, a similar trigger identification implementation is present in the open source code for ESL \cite{esl_code}.

\subsection{Camera settings}

We are interested only in positive events, as we have designed our approach to work with events triggered by the laser passing over camera pixels. Thus, we set the bias that controls the threshold for negative changes \textit{diff\_off} to its minimum value, and filter out any remaining negative events from the stream. We set the deadtime \textit{refr} and the high pass filter \textit{hpf} to their lowest allowed values, so the camera produces as many events from neighboring rows as possible. When setting up a demonstrator under new lighting conditions, we set \textit{diff\_on} to the minimum possible value where we do not see any noise events outside the projected image.

\subsection{Event filtering}
\label{sec:event_filtering}

When creating time maps, only one value can be stored at each coordinate. If there are multiple events for the same $(x,y)$ within one frame, the duplicates get filtered out implicitly. As our method does not depend on creating a time map, we may keep multiple events per coordinate, and process them all. This is a trade-off: Keeping all events allows the resulting point cloud to be more dense, but it will also have visibly more noise in the depth, making thin structures grow thicker. Nevertheless, we see the possibility of assigning a 3D point to each event as a great benefit of our method. In testing, we found that more than 60 \% of all events may be dropped by the coordinate filter. In future work, the higher information density by keeping those could be used to optimize the quality of the measurements and to model the behavior of the projector and camera more accurately.

\section{Evaluation and Discussion}

\subsection{Time map calibration}

\begin{figure}[t]
  \centering
   \includegraphics[width=\linewidth]{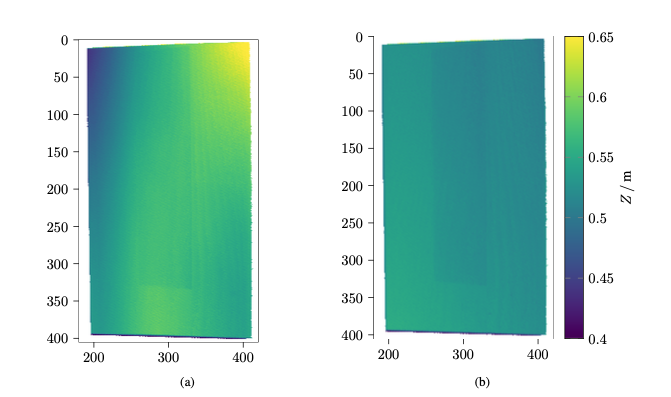}

   \caption{Comparison for a projection straight onto a plane, using calibrated and uncalibrated projector time maps. Ideally, we would see a depth map that shares the same values over the whole region. Figure (a) depicts the estimated depth map without projector time map calibration, displaying noticeable distortions in the upper corners of the simple plane shape. Figure (b) presents the depth map generated from the same events, but with a pre-calibrated projector time map applied. This demonstrates, how the time map calibration counteracts the temporal non-linearities of the projector well. 
}
  \label{fig:time_map_calib}
\end{figure}

The evaluation of the time map calibration, as illustrated in Figure \ref{fig:time_map_calib}, demonstrates that the calibration methodology is fundamentally effective. The depth measurements presented in the figure validate the underlying principle of the time map calibration process. To obtain a comprehensive understanding of the performance, a thorough qualitative evaluation is required in future studies. Additionally, integrating intrinsic temporal calibration of the projector as a component of the calibration process could potentially enhance the precision and accuracy further, leading to an overall more effective system.

Two noteworthy observations have emerged from the time map calibration experiments. First, the calibration exhibited stability across various projector and camera lens  calibrations. The once-calibrated time map for our projector could be effectively reapplied months later, yielding accurate results without requiring recalibration. Second, we observed that different projectors exhibit varying degrees of nonlinear temporal behavior. In the data recorded for the ESL \cite{Muglikar2021}, the projector displayed a comparatively milder nonlinear behavior, resulting in significantly flatter planes even without time map calibration.

\subsection{Comparison to the state of the art}

\begin{table*}[t!]
\setlength\tabcolsep{2 pt}
\footnotesize
\begin{center}
  \begin{tabular}{l cc cc cc cc cc cc cc cc cc}
  \toprule
    Scene & \multicolumn{2}{c}{David} & \multicolumn{2}{c}{Heart} & \multicolumn{2}{c}{Book-Duck} & \multicolumn{2}{c}{Plant} & \multicolumn{2}{c}{City of Lights} & \multicolumn{2}{c}{Cycle} & \multicolumn{2}{c}{Room} & \multicolumn{2}{c}{Desk-chair} & \multicolumn{2}{c}{Desk-books} \\
    \cmidrule(lr){2-3}\cmidrule(lr){4-5}\cmidrule(lr){6-7}\cmidrule(lr){8-9}\cmidrule(lr){10-11}\cmidrule(lr){12-13}\cmidrule(lr){14-15}\cmidrule(lr){16-17} \cmidrule(lr){18-19}
    Metrics &  FR$\uparrow$ & RMSE$\downarrow$ & FR$\uparrow$& RMSE$\downarrow$ & FR$\uparrow$ & RMSE$\downarrow$ & FR$\uparrow$ & RMSE$\downarrow$ & FR$\uparrow$ & RMSE$\downarrow$ & FR$\uparrow$ & RMSE$\downarrow$ & FR$\uparrow$ & RMSE$\downarrow$ & FR$\uparrow$ & RMSE$\downarrow$ & FR$\uparrow$ & RMSE$\downarrow$ \\
    \midrule
MC3D & 0.06 & 1.28  & 0.06 & 1.33  & 0.08 & 2.32  & 0.06 & 4.19  & 0.06 & 7.47  & 0.04 & 6.46  & 0.07 & 4.86  & 0.07 & 2.98  & 0.09 & 1.77 \\
MC3D-1s & 0.44 & 0.76  & 0.45 & 0.74  & 0.54 & 1.01  & 0.39 & 3.31  & 0.41 & 4.36  & 0.29 & 3.58  & 0.31 & 10.33  & 0.38 & 4.39  & 0.6 & 2.48 \\      
ESL-init & \textbf{0.98} & \textbf{0.21}  & \textbf{0.98} & \textbf{0.2}  & \textbf{0.91} & \textbf{0.3}  & \textbf{0.88} & \textbf{0.44}  & \textbf{0.86} & \textbf{0.69}  & \textbf{0.9} & \textbf{0.56}  & \textbf{0.81} & \textbf{0.53}  & \textbf{0.93} & \textbf{0.26}  & \textbf{0.98} & \textbf{0.19} \\
X-maps (ours) & \textbf{0.97} & \textbf{0.21}  & \textbf{0.98} & \textbf{0.2}  & \textbf{0.91} & \textbf{0.31}  & \textbf{0.88} & \textbf{0.45}  & \textbf{0.85} & 0.72  & \textbf{0.89} & \textbf{0.57}  & 0.74 & \textbf{0.53}  & 0.88 & \textbf{0.25}  & 0.91 & \textbf{0.19} \\    
    \bottomrule
  \end{tabular}
\end{center}
  \caption{\textbf{Evaluation with static scenes}: Fill rate (FR, depth map completion) and RMSE (cm) are measured with respect to a ground truth (optimized ESL). If the best and second best approaches differ by only $\SI{0.01}{\cm}$ or $\SI{0.01}{\percent}$, both are highlighted.}
  \label{tab:stateoftheart}
\end{table*}

We compare our method to the state of the art approach ESL \cite{Muglikar2021} and use the static scenes of the public dataset provided by them. We do not know of a dataset to provide event data for projected images together with a ground truth depth. In ESL, results from a time-averaged MC3D calculation are used as a baseline to compare against. We found that the refined ESL results (window size $W = 7$ and denoised) capture the geometry more cleanly than MC3D. Thus, we consider the ESL results as the best approximation of scene depth, and use them as the improved baseline. The event camera and the laser projector were synchronized via an external sync jack, so it is assumed that all triggers are perfect. The frequency of the images is $\SI{60}{\hertz}$. Our time map calibration cannot be applied on the dataset, as it does not contain an image of a flat surface. Thus, a linear temporal behaviour of the projector has to be assumed.

\paragraph{Evaluation metrics.} We list the root mean square error (RMSE), the Euclidean distance between estimates and baseline (in all valid regions in both images) in \SI{}{\cm}, and the fill-rate (completeness). The fill-rate measures the percentage of ground truth points for which the distance to the estimation is smaller than a threshold. The threshold is $\SI{1}{\percent}$ of the average scene depth. The metrics are calculated for each provided time map, and, additionally, the average value is listed. Since the scenes are static, the average values differ very little from individual frame measurements.

\paragraph{Quantitative results.} 
Table \ref{tab:stateoftheart} shows the quantitative results of the depth estimation. We compare against \textit{MC3D}~\footnote{The MC3D and ESL (CPU) implementation we used in the evaluation is part of the open source implementation of ESL\cite{esl_code}} and the initialization step of ESL (\textit{ESL-init}), which uses a row-wise disparity matching and no further optimization. Our X-maps system provides results very similar to \mbox{ESL-init}, with an RMSE difference of $\SI{0.7}{\mm}$ at maximum and almost equal fill rate. This demonstrates that our fast X-map lookup provides depth at the same quality as the exhaustive disparity search. The small differences can be explained by the different quantizations (in the $x$-axis for the time map, and the $t$-axis for the X-map), the rejection of values when creating the reference projector X-map (see Section \ref{sec:projector-reference-x-map}) and interpolation and border handling errors in the remapping steps. In comparison to MC3D, we see that X-maps provides depth maps with much higher fill rate and lower RMSE. MC3D is not able to capture full frames with a frequency of $\SI{60}{\hertz}$. If the MC3D measurements are averaged over a period of 1 second (60 frames) in \mbox{MC3D-1s}, the depth maps become more dense, but still differ a lot from the smoothed ESL depth maps.

In this experiment, we use static scenes, but due to the high temporal resolution of event cameras and the fact that both ESL and our method do not use temporal filtering, the results can be transferred to scenes with movements. The increased accuracy in dynamic scenes and robustness of event-based sensing compared to structured light sensors such as RealSense \cite{IntelRealSense} was performed in \cite{Muglikar2021} and is also applicable to our method.

\begin{figure}[t]
  \centering
   \includegraphics[width=\linewidth]{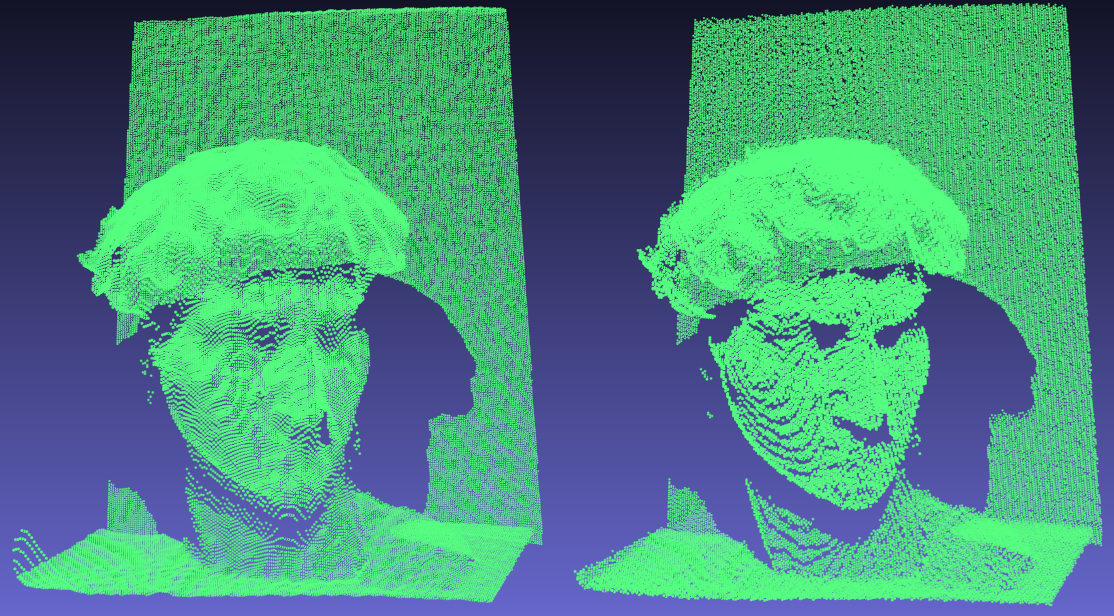}
   \caption{Qualitative comparison of reconstructed point clouds using ESL with depth map optimization (\textit{W = 7}) (left) and X-maps (right). The smoothing of the ESL point cloud is computationally expensive. Our point cloud contains slightly more noise, but can be determined without computational effort and still captures geometric details.}
   \label{fig:pointcloud_comparison}
\end{figure}

\paragraph{Qualitative result and runtime evaluation.} Figure \ref{fig:pointcloud_comparison} shows a point cloud comparison for the \textit{David} scene. We see that the point cloud acquired with our method can preserve the geometric details of the scene very well. The ESL point cloud is smoother, visible especially in flat regions (e.g.~the wall) and slightly better filled in some regions (e.g.~at the eyes), but the calculation is also computationally very expensive and time consuming.  

Table \ref{tab:runtime_comparison} provides a detailed timing evaluation. We perform the tests on a system equipped with an AMD Ryzen Threadripper PRO 5955WX CPU and an RTX 4090 GPU, running Python 3.8. On a single CPU, one depth map calculation with X-maps takes $\SI{2.67}{\ms}$ on average, while one iteration of ESL (with window size $W = 7$ and denoising) took over 100 seconds, making it over $10000\times$ slower than X-maps. 
 A faster implementation of ESL is possible, but we believe that a GPU implementation is necessary for row-by-row disparity search and depth optimization to run in real-time. To demonstrate this, we create an optimized CUDA implementation of ESL-init which required $\SI{18.99}{\ms}$ on GPU, still $7\times$ slower than X-maps.

\begin{table}[ht]
\setlength\tabcolsep{2 pt}
\footnotesize
\begin{center}
\begin{tabular}{lccc}
\toprule
\textbf{Method} & Runtime (abs.)  & Runtime (rel.)  \\
\midrule
ESL (CPU) & 174.68 s $(\pm 26.97)$ & $> 10000\times$\\
ESL-init (CPU) & 11.87 s $(\pm 2.13)$ & $> 1000\times$\\
ESL-init (CUDA) & 18.99 ms $(\pm 0.88)$ & $7.12\times$\\
X-maps (ours, CPU) & \textbf{2.67 ms} $(\pm 0.31)$ & $\mathbf{1.0\times}$  \\
\bottomrule
\end{tabular}
\end{center}
\caption{\textbf{Runtime comparison}: For ESL-init, we additionally measure the runtime of our own CUDA-based implementation. We calculate the timing for each static scene and list the average time and the standard deviation.}
\label{tab:runtime_comparison}
\end{table}

\subsection{Spatial Augmented Reality Example}

With our method, we have built a live demonstrator, that computes the depth of the scene, and projects the depth values back into the scene in real-time (Figure \ref{fig:comb_scene_shots}). Our method does not depend on a certain kind of pattern (like a noise or binary pattern), and, crucially, supports projecting with all available pixels of the projector. Thus, we can display any image content onto the scene, while still estimating depth information. It is also possible to project sparse images, as our method computes depth for events, and is not dependent on the image being filled. Note that if the scene gets too sparse, the frame trigger algorithm described in Section \ref{sec:frame_triggers} may stop working correctly. Very sparse scenes would require a change in the algorithm that can work with the projected scene as an input to find the frame beginning and end.

In the future, we want to create more complex experiences, by combining our depth estimation method with object detection and tracking, so that we are able to overlay real-world objects with information or patterns.

\section{Conclusions, Limitations \& Future Work}

We have presented a method of directly computing the depth of events, triggered by a small laser projector. After transforming the projector time map into a rectified \mbox{X-map}, the depth can directly be retrieved from a lookup table, while accounting for non-linear temporal behavior of the scanner. Thus, the method allows for the creation of high-resolution point clouds at $\SI{60}{\hertz}$, or higher. This enables exciting use cases in spatial augmented reality. The projector is cheap, and little computational budget is required.

Our system is limited by the noise of the event camera sensor and its readout characteristics, which introduces errors in the timestamps. Improvements in sensor resolution, noise reduction and readout speed will directly benefit our approach, without needing any changes.

We have chosen not to apply further noise filtering to the data in the image domain, keeping depth information from all events in the resulting scene. This yields sharp borders at depth discontinuities, no erroneous interpolation, and even very small objects stay visible. We trade this off for a reduced smoothness in the depth map, which we consider sufficient for spatial augmented reality use cases. We have evaluated our results by showing that our approach provides similar depth precision to earlier work on a public dataset. 

In future work, it might be interesting to closer investigate the row-wise behaviour of the laser scanner, in order to provide a more exact time map calibration. Finally, in order to allow projection onto arbitrary (potentially moving) objects for different applications, the projection of partial frames with any contour should be investigated.

\section*{Acknowledgements}

We would like to thank Prophesee for providing us with an event sensor evaluation kit, which we used in our demonstrator.
This work has been funded by the German Federal Ministry of Economic Affairs and Climate Actions under grant number 01MT20001D (Gemimeg).

{\small
\bibliographystyle{ieee_fullname}
\bibliography{egbib_wm}
}

\end{document}